\documentclass[11pt,a4paper]{article}
\usepackage[hyperref]{emnlp-ijcnlp-2019}
\usepackage{times}
\usepackage{latexsym}

\usepackage{url}

\aclfinalcopy 

\usepackage[normalem]{ulem}
\usepackage{soul}
\usepackage{enumitem}
\usepackage{adjustbox}
\usepackage{amsmath}
\usepackage{amssymb}
\usepackage{multirow}
\usepackage{array}
\usepackage{graphicx}
\usepackage{comment}
\usepackage{booktabs}
\usepackage{subcaption}
\usepackage{caption}
\usepackage{xcolor,colortbl} 
\usepackage{multirow}
\usepackage{makecell}

\DeclareMathOperator*{\argmax}{arg\,max}

\title{Improving Question Generation With to the Point Context}
\author{Jingjing Li\textsuperscript{1}\thanks{\quad These two authors contributed equally.} ~~~~~~ Yifan Gao\textsuperscript{1}\footnotemark[1] ~~~~~~ Lidong Bing\textsuperscript{2}
~~~~~~ Irwin King\textsuperscript{1} ~~~~~~ Michael R. Lyu\textsuperscript{1}  \\
	{\textsuperscript{1} Department of Computer Science and Engineering, }\\
	{The Chinese University of Hong Kong, Shatin, N.T., Hong Kong}\\
	{\textsuperscript{2} R\&D Center Singapore, Machine Intelligence Technology}\\ {Alibaba DAMO Academy}\\
    { \textsuperscript{1}\{lijj, yfgao, king, lyu\}@cse.cuhk.edu.hk}
	{ \textsuperscript{2}l.bing@alibaba-inc.com}
}

\date{}

\begin{document}
\maketitle
\begin{abstract}
Question generation (QG) is the task of generating a question from a reference sentence and a specified answer within the sentence. 
A major challenge in QG is to identify answer-relevant context words to finish the declarative-to-interrogative sentence transformation.
Existing sequence-to-sequence neural models achieve this goal by proximity-based answer position encoding under the intuition that neighboring words of answers are of high possibility to be answer-relevant.
However, such intuition may not apply to all cases especially for sentences with complex answer-relevant relations.
Consequently, the performance of these models drops sharply when the relative distance between the answer fragment and other non-stop sentence words that also appear in the ground truth question increases.
To address this issue, we propose a method to jointly model the unstructured sentence and the structured answer-relevant relation (extracted from the sentence in advance) for question generation.
Specifically, the structured answer-relevant relation acts as the to the point context and it thus naturally helps keep the generated question to the point, while the unstructured sentence provides the full information.
Extensive experiments show that to the point context helps our question generation model achieve significant improvements on several automatic evaluation metrics.
Furthermore, our model is capable of generating diverse questions for a sentence which conveys multiple relations of its answer fragment.
\end{abstract}

\section{Introduction} \label{sec.intro}

\begin{figure}[t!]
\small
\begin{tabular}{p{0.9\columnwidth}}
\hline\hline
\textbf{Sentence}: The daily mean temperature in January, the area's coldest month, is 32.6 $^\circ$F (\hl{0.3 $^\circ$C}); however, temperatures usually drop to 10 $^\circ$F (-12 $^\circ$C) several times per winter and reach 50 $^\circ$F (10 $^\circ$C) several days each winter month. \\
\textbf{Reference Question}: What is New York City 's daily January mean temperature in degrees celsius ? \\
\textbf{Baseline Prediction}: What is the coldest temperature in Celsius ? \\
\textbf{Structured Answer-relevant Relation}: (The daily mean temperature in January; is; 32.6 $^\circ$F (0.3 $^\circ$C))\\
\hline\hline                
\end{tabular}
\caption{An example SQuAD question with the baseline's prediction. The answer (``0.3 $^\circ$C'') is highlighted.}
\label{fig.example}
\end{figure}

Question Generation (QG) is the task of automatically creating questions from a range of inputs, such as natural language text \cite{Heilman2010GoodQS}, knowledge base \cite{Serban2016GeneratingFQ} and image \cite{Mostafazadeh2016GeneratingNQ}.
QG is an increasingly important area in NLP with various application scenarios such as intelligence tutor systems, open-domain chatbots and question answering dataset construction.
In this paper, we focus on question generation from reading comprehension materials like SQuAD \cite{Rajpurkar2016SQuAD10}.
As shown in Figure \ref{fig.example}, given a sentence in the reading comprehension paragraph and the text fragment (i.e., the answer) that we want to ask about, we aim to generate a question that is asked about the specified answer.

\begin{table*}[t!]
    \centering
    {
    \resizebox{0.6\textwidth}{!}{
    \begin{tabular}{c|cccccc} 
    \Xhline{3\arrayrulewidth}
    Distance & B1 & B2 & B3 & B4 & MET & R-L     \\ 
    \hline
    0$\sim$10~~ (72.8\% of \#)   & 45.25	& 30.31 & 22.06	& 16.54	& 21.54	& 46.26    \\ 
    $>$10~~ (27.2\% of \#)  & 35.67 &	21.72	&	14.82	&	10.46	&	16.72	&	37.63    \\ 
    \Xhline{3\arrayrulewidth}
    \end{tabular}}
    }
    \caption{Performance for the average relative distance between the answer fragment and other non-stop sentence words that also appear in the ground truth question. (Bn: BLEU-n, MET: METEOR, R-L: ROUGE-L)}
    \label{tab:distance}
\end{table*}

Question generation for reading comprehension is firstly formalized as a declarative-to-interrogative sentence transformation problem with predefined rules or templates \cite{Mitkov2003ComputeraidedGO,Heilman2010GoodQS}.
With the rise of neural models, \newcite{Du2017LearningTA} propose to model this task under the sequence-to-sequence (Seq2Seq) learning framework \cite{Sutskever2014SequenceTS} with attention mechanism \cite{Luong2015EffectiveAT}.
However, question generation is a one-to-many sequence generation problem, i.e., several aspects can be asked given a sentence.
\newcite{Zhou2017NeuralQG} propose the answer-aware question generation setting which assumes the answer, a contiguous span inside the input sentence, is already known before question generation.
To capture answer-relevant words in the sentence, they adopt a BIO tagging scheme to incorporate the answer position embedding in Seq2Seq learning. 
Furthermore, \newcite{Sun2018AnswerfocusedAP} propose that tokens close to the answer fragments are more likely to be answer-relevant.
Therefore, they explicitly encode the relative distance between sentence words and the answer via position embedding and position-aware attention.

Although existing proximity-based answer-aware approaches achieve reasonable performance, we argue that such intuition may not apply to all cases especially for sentences with complex structure. 
For example, Figure \ref{fig.example} shows such an example where those approaches fail.
This sentence contains a few facts and due to the parenthesis (i.e. ``the area's coldest month''), some facts intertwine: ``The daily mean temperature in January is 0.3$^\circ$C'' and ``January is the area's coldest month''. From the question generated by a proximity-based answer-aware baseline, we find that it wrongly uses the word ``coldest'' but misses the correct word ``mean'' because ``coldest'' has a shorter distance to the answer ``0.3$^\circ$C''.

In summary, their intuition that ``the neighboring words of the answer are more likely to be answer-relevant and have a higher chance to be used in the question'' is not reliable. 
To quantitatively show this drawback of these models, we implement the approach proposed by \newcite{Sun2018AnswerfocusedAP} and analyze its performance under different relative distances between the answer and other non-stop sentence words that also appear in the ground truth question.
The results are shown in Table \ref{tab:distance}. We find that the performance drops at most 36\% when the relative distance increases from ``$0\sim10$'' to ``$>10$''. In other words, when the useful context is located far away from the answer, current proximity-based answer-aware approaches will become less effective, since they overly emphasize neighboring words of the answer.

To address this issue, we extract the structured answer-relevant relations from sentences and propose a method to jointly model such structured relation and the unstructured sentence for question generation.
The structured answer-relevant relation is likely to be to the point context and thus can help keep the generated question to the point.
For example, Figure \ref{fig.example} shows our framework can extract the right answer-relevant relation (``The daily mean temperature in January'', ``is'', ``32.6$^\circ$F (0.3$^\circ$C)'') among multiple facts.
With the help of such structured information, our model is less likely to be confused by sentences with a complex structure. 
Specifically, we firstly extract multiple relations with an off-the-shelf Open Information Extraction (OpenIE) toolbox \cite{Saha2018OpenIE}, then we select the relation that is most relevant to the answer with carefully designed heuristic rules. 

Nevertheless, it is challenging to train a model to effectively utilize both the unstructured sentence and the structured answer-relevant relation because both of them could be noisy: the unstructured sentence may contain multiple facts which are irrelevant to the target question, while the limitation of the OpenIE tool may produce less accurate extracted relations.
To explore their advantages simultaneously and avoid the drawbacks, we design a gated attention mechanism and a dual copy mechanism based on the encoder-decoder framework, where the former learns to control the information flow between the unstructured and structured inputs, while the latter learns to copy words from two sources to maintain the informativeness and faithfulness of generated questions.

In the evaluations on the SQuAD dataset, our system achieves significant and consistent improvement as compared to all baseline methods.
In particular, we demonstrate that the improvement is more significant with a larger relative distance between the answer and other non-stop sentence words that also appear in the ground truth question.
Furthermore, our model is capable of generating diverse questions for a single sentence-answer pair where the sentence conveys multiple relations of its answer fragment.


\section{Framework Description}
In this section, we first introduce the task definition and our protocol to extract structured answer-relevant relations.
Then we formalize the task under the encoder-decoder framework with gated attention and dual copy mechanism.

\subsection{Problem Definition}
We formalize our task as an answer-aware Question Generation (QG) problem \cite{Zhao2018ParagraphlevelNQ}, which assumes answer phrases are given before generating questions. Moreover, answer phrases are shown as text fragments in passages.
Formally, given the sentence $S$, the answer $A$, and the answer-relevant relation $M$, the task of QG aims to find the best question $\overline{Q}$ such that,
\begin{align}
    \overline{Q} = \argmax_{Q} \text{Prob(}Q|S, A, M\text{)},
\end{align}
where $A$ is a contiguous span inside $S$.

\begin{figure}[t!]
\small
\begin{tabular}{p{0.95\columnwidth}}
\hline\hline
\textbf{Sentence}: Beyonc\'{e} received critical acclaim and commercial success, selling \hl{one million} digital copies worldwide in six days; The New York Times noted the album's unconventional, unexpected release as significant. \\
\textbf{N-ary Relations}: \\
0.85 \uwave{(Beyonc\'{e}; received commercial success selling; one million digital copies worldwide; in six days)}\\
0.92 (The New York Times; noted; the album's unconventional, unexpected release as significant)\\
\hline
\textbf{Sentence}: The daily mean temperature in January, the area's coldest month, is 32.6 $^\circ$F (\hl{0.3 $^\circ$C}); however, temperatures usually drop to 10 $^\circ$F (-12 $^\circ$C) several times per winter and reach 50 $^\circ$F (10 $^\circ$C) several days each winter month. \\
\textbf{N-ary Relations}: \\
0.95 \uwave{(The daily mean temperature in January; is; 32.6 $^\circ$F (0.3 $^\circ$C))}\\
0.94 (temperatures; drop; to 10 $^\circ$F (−12 $^\circ$C); several times per winter; usually)\\
0.90 (temperatures; reach; 50 $^\circ$F)\\
\hline\hline                
\end{tabular}
\caption{Examples for n-ary extractions from sentences using OpenIE. Confidence scores are shown at the beginning of each relation. Answers are highlighted in sentences. Waved relations are selected according to our criteria in Section \ref{sec.model.openie}. }
\label{fig.example_openie}
\end{figure}

\subsection{Answer-relevant Relation Extraction} \label{sec.model.openie}
We utilize an off-the-shelf toolbox of OpenIE \footnote{http://openie.allenai.org/} to the derive structured answer-relevant relations from sentences as to the point contexts.
Relations extracted by OpenIE can be represented either in a triple format or in an n-ary format with several secondary arguments, and we employ the latter to keep the extractions as informative as possible and avoid extracting too many similar relations in different granularities from one sentence.
We join all arguments in the extracted n-ary relation into a sequence as our to the point context.
Figure \ref{fig.example_openie} shows n-ary relations extracted from OpenIE.
As we can see, OpenIE extracts multiple relations for complex sentences. 
Here we select the most informative relation according to three criteria in the order of descending importance: (1) having the maximal number of overlapped tokens between the answer and the relation; (2) being assigned the highest confidence score by OpenIE; (3) containing maximum non-stop words.
As shown in Figure \ref{fig.example_openie}, our criteria can select answer-relevant relations (waved in Figure \ref{fig.example_openie}), which is especially useful for sentences with extraneous information. 
In rare cases, OpenIE cannot extract any relation, we treat the sentence itself as the to the point context.

Table \ref{tab:openie_stat} shows some statistics to verify the intuition that the extracted relations can serve as more to the point context.
We find that the tokens in relations are 61\% more likely to be used in the target question than the tokens in sentences, and thus they are more to the point.
On the other hand, on average the sentences contain one more question token than the relations (1.86 v.s. 2.87). 
Therefore, it is still necessary to take the original sentence into account to generate a more accurate question.

\begin{table}[t!]
    \centering
    {
    \resizebox{1.0\columnwidth}{!}{
    \begin{tabular}{l|cc} 
    \Xhline{2\arrayrulewidth}
     & Sentence & Answer-relevant Relation    \\ 
    \hline
    Avg. length   & 32.46	& 13.04     \\ 
    
    \# overlapped words  & ~~2.87 &	~~1.86\\ 
    
    Copy ratio & ~~~~~8.85\% & ~~~14.26\% \\
    \Xhline{2\arrayrulewidth}
    \end{tabular}}
    }
    \caption{Comparisons between sentences and answer-relevant relations. Overlapped words are those non-stop tokens co-occurring in the source (sentence/relation) and the target question. Copy  ratio means the proportion of source tokens that are used in the question.}
    \label{tab:openie_stat}
\end{table}

\subsection{Our Proposed Model}
\begin{figure*}[t!]
\centering
\includegraphics[width=1.0\textwidth]{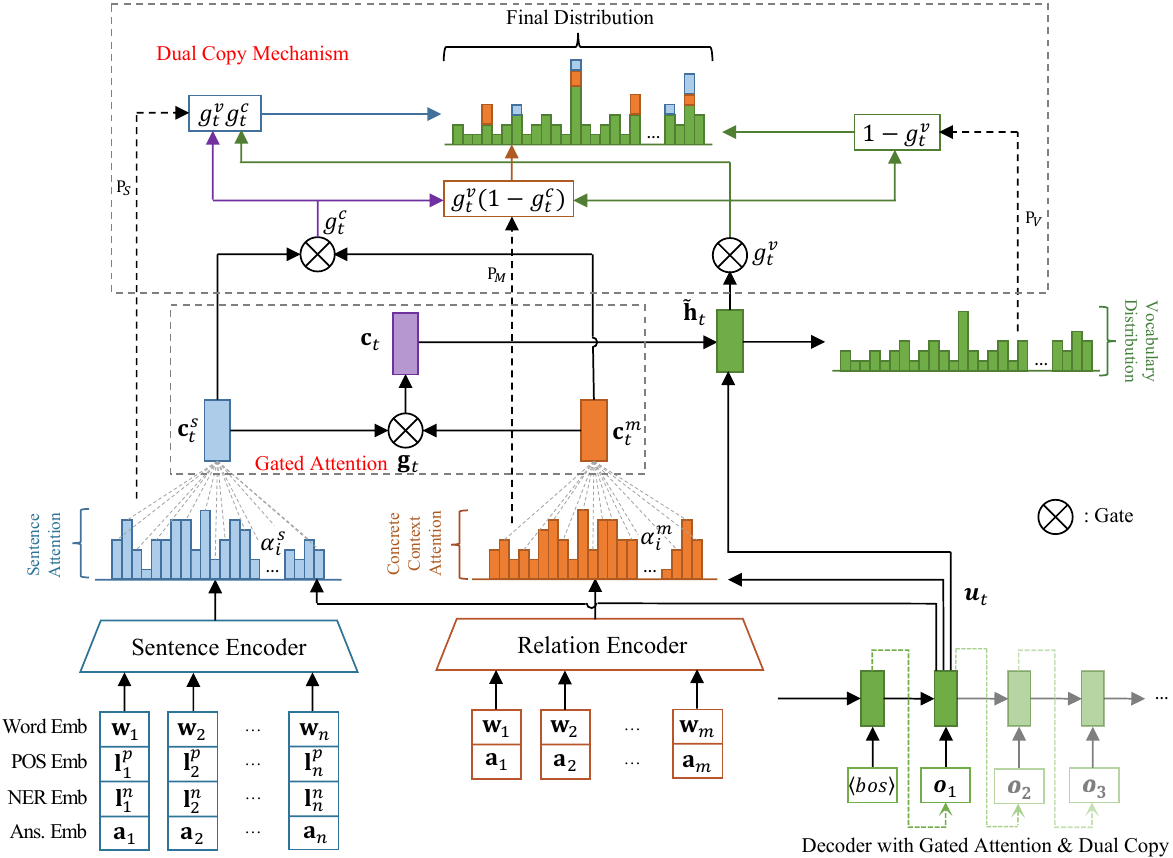}
\caption{The framework of our proposed model. \textit{(Best viewed in color)} }
\label{figure:framework}
\end{figure*}

\paragraph{Overview.}
As shown in Figure \ref{figure:framework}, our framework consists offour components (1) Sentence Encoder and Relation Encoder, (2) Decoder, (3) Gated Attention Mechanism and (4) Dual Copy Mechanism.
The sentence encoder and relation encoder encode the unstructured sentence and the structured answer-relevant relation, respectively.
To select and combine the source information from the two encoders, a gated attention mechanism is employed to jointly attend both contextualized information sources, and a dual copy mechanism copies words from either the sentence or the relation.

\paragraph{Answer-aware Encoder.}
We employ two encoders to integrate information from the unstructured sentence $S$ and the answer-relevant relation $M$ separately. 
Sentence encoder takes in feature-enriched embeddings including word embeddings $\mathbf w$, linguistic embeddings $\mathbf l$ and answer position embeddings $\mathbf a$.
We follow \cite{Zhou2017NeuralQG} to transform POS and NER tags into continuous representation ($\mathbf{l}^p$ and $\mathbf{l}^n$) and adopt a BIO labelling scheme to derive the answer position embedding (B: the first token of the answer, I: tokens within the answer fragment except the first one, O: tokens outside of the answer fragment).
For each word $w_i$ in the sentence $S$, we simply concatenate all features as input: $\mathbf{x}_i^s= [\mathbf{w}_i; \mathbf{l}^p_i; \mathbf{l}^n_i; \mathbf{a}_i]$.
Here $[\mathbf a;\mathbf b]$ denotes the concatenation of vectors $\mathbf a$ and $\mathbf b$.

We use bidirectional LSTMs to encode the sentence $(\mathbf{x}_1^s, \mathbf{x}_2^s, ..., \mathbf{x}_n^s)$ to get a contextualized representation for each token:
\begin{align}\label{eqn.enc_sent}
\resizebox{0.95\hsize}{!}{
$\overrightarrow{\mathbf{h}}_i^s = \overrightarrow{\text{{LSTM}}}(\overrightarrow{\mathbf{h}}_{i-1}^s, \mathbf{x}_i^s),~
\overleftarrow{\mathbf{h}}_i^s = \overleftarrow{\text{{LSTM}}}(\overleftarrow{\mathbf{h}}_{i+1}^s, \mathbf{x}_i^s),\nonumber$}
\end{align}
where $\overrightarrow{\mathbf{h}}^{s}_i$ and $\overleftarrow{\mathbf{h}}^{s}_i$ are the hidden states at the  $i$-th time step of the forward and the backward LSTMs. 
The output state of the sentence encoder is the concatenation of forward and backward hidden states: $\mathbf{h}^{s}_i=[\overrightarrow{\mathbf{h}}^{s}_i;\overleftarrow{\mathbf{h}}^{s}_i]$. 
The contextualized representation of the sentence is $(\mathbf{h}^{s}_1, \mathbf{h}^{s}_2, ..., \mathbf{h}^{s}_n)$.

For the relation encoder, we firstly join all items in the n-ary relation $M$ into a sequence. Then we only take answer position embedding as an extra feature for the sequence: $\mathbf{x}_i^m= [\mathbf{w}_i; \mathbf{a}_i]$. Similarly, we take another bidirectional LSTMs to encode the relation sequence and derive the corresponding contextualized representation $(\mathbf{h}^{m}_1, \mathbf{h}^{m}_2, ..., \mathbf{h}^{m}_n)$.

\paragraph{Decoder.}
We use an LSTM as the decoder to generate the question. 
The decoder predicts the word probability distribution at each decoding timestep to generate the question. At the \textit{t}-th timestep, it reads the word embedding $\mathbf{w}_{t}$ and the hidden state $\mathbf{u}_{t-1}$ of the previous timestep to generate the current hidden state:
\begin{align}
    \mathbf u_t = \text{LSTM}(\mathbf u_{t-1}, \mathbf w_{t}).
\end{align}

\paragraph{Gated Attention Mechanism.}
We design a gated attention mechanism to jointly attend the sentence representation and the relation representation. 
For sentence representation $(\mathbf{h}^{s}_1, \mathbf{h}^{s}_2, ..., \mathbf{h}^{s}_n)$, we employ the \newcite{Luong2015EffectiveAT}'s attention mechanism to obtain the sentence context vector $\mathbf{c}^s_t$,
\begin{align}
    a^s_{t,i} = \frac{\text{exp}(\mathbf{u}_t^\top \mathbf{W}_a \mathbf{h}^s_i)}{\sum_{j} \text{exp}(\mathbf{u}_t^\top \mathbf{W}_a \mathbf{h}^s_j)},  ~
    \mathbf{c}^s_t = \sum_i a^s_{t,i}\mathbf{h}^s_i,\nonumber
\end{align}
where $\mathbf{W}_a$ is a trainable weight.
Similarly, we obtain the vector $\mathbf{c}^m_t$ from the relation representation $(\mathbf{h}^{m}_1, \mathbf{h}^{m}_2, ..., \mathbf{h}^{m}_n)$.
To jointly model the sentence and the relation, a gating mechanism is designed to control the information flow from two sources:
\begin{align}
    \mathbf{g}_t &= \text{sigmoid}(\mathbf{W}_g[\mathbf{c}^s_t; \mathbf{c}^m_t]), \\
    \mathbf{c}_t &= \mathbf{g}_t \odot \mathbf{c}^s_t + (\mathbf 1-\mathbf{g}_t) \odot \mathbf{c}^m_t , \\
    \tilde{\mathbf{h}}_t  &= \text{tanh}(\mathbf{W}_{h} [\mathbf{u}_t;\mathbf{c}_t]),
\end{align}
where $\odot$ represents element-wise dot production and $\mathbf{W}_g, \mathbf{W}_h$ are trainable weights. 
Finally, the predicted probability distribution over the vocabulary $V$ is computed as:
\begin{align}\label{dec.out}
\text{P}_{V} = \text{softmax}(\mathbf{W}_V \tilde{\mathbf{h}}_{t} + \mathbf{b}_V),
\end{align}
where $\mathbf{W}_V$ and $\mathbf{b}_V$ are parameters.

\paragraph{Dual Copy Mechanism.}
To deal with the rare and unknown words, the decoder applies the pointing method \cite{See2017GetTT,Gu2016IncorporatingCM,Glehre2016PointingTU} to allow copying a token from the input sentence at the $t$-th decoding step. 
We reuse the attention score $\mathbf{\alpha}_{t}^s$ and $\mathbf{\alpha}_{t}^m$ to derive the copy probability over two source inputs:
\begin{align}
    \text{P}_S(w) &= \sum_{i: w_i=w}{\alpha}^s_{t,i},~
    \text{P}_M(w) &= \sum_{i: w_i=w}{\alpha}^m_{t,i}.\nonumber
\end{align}
Different from the standard pointing method, we design a dual copy mechanism to copy from two sources with two gates.
The first gate is designed for determining copy tokens from two sources of inputs or generate next word from $P_V$, which is computed as $g^v_t = \text{sigmoid}(\mathbf{w}^v_g \tilde{\mathbf{h}}_t + b^v_g)$.
The second gate takes charge of selecting the source (sentence or relation) to copy from, which is computed as $g^c_t = \text{sigmoid}(\mathbf{w}^c_g [\mathbf{c}_t^s;\mathbf{c}_t^m] + b^c_g)$.
Finally, we combine all probabilities $P_V$, $P_S$ and $P_M$ through two soft gates $g^v_t$ and $g^c_t$. The probability of predicting $w$ as the $t$-th token of the question is:
\begin{align}
\text{P}(w) &= (1-g^v_t) \text{P}_V(w) \\
&+ g^v_t g^c_t \text{P}_S(w) + g^v_t (1-g^c_t) \text{P}_M(w) \nonumber.
\end{align}

\paragraph{Training and Inference.}
Given the answer $A$, sentence $S$ and relation $M$, the training objective is to minimize the negative log-likelihood with regard to all parameters:
\begin{align}
    \mathcal{L} = - \sum_{Q\in \mathcal \{Q\}} log ~ \text{P}(Q|A,S,M;\theta),
\end{align}
where $\mathcal \{Q\}$ is the set of all training instances,  $\theta$ denotes model parameters and $\text{log} P(Q|A,S,M;\theta)$ is the conditional log-likelihood of $Q$. 

In testing, our model targets to generate a question $Q$ by maximizing:
\begin{align}
    \overline{Q} = \arg\max_{Q} log ~\text{P}(Q|A,S,M;\theta).
\end{align}

\section{Experimental Setting}
\subsection{Dataset \& Metrics}\label{sec:dataset}
\begin{table}[!t]
    \centering
    \resizebox{0.9\columnwidth}{!}
    {\small
    \begin{tabular}{l| c c}
    \Xhline{2\arrayrulewidth}
    & Du Split & Zhou Split \\ \hline
    \# pairs (Train)  & 74689    & 86635   \\
    \# pairs (Dev)    & 10427  & 8965     \\ 
     \# pairs (Test)    & 11609  & 8964     \\ 
    Sentence avg. tokens  & 32.56  & 32.72     \\ 
    Question avg. tokens  & 11.42  & 11.31     \\ 
    \Xhline{2\arrayrulewidth}
    \end{tabular}}
    \caption{Dataset statistics on Du Split \cite{Du2017LearningTA} and Zhou Split \cite{Zhou2017NeuralQG}.}
    \label{tab.dataset}
\end{table}

We conduct experiments on the SQuAD dataset \cite{Rajpurkar2016SQuAD10}.
It contains 536 Wikipedia articles and 100k crowd-sourced question-answer pairs. The questions are written by crowd-workers and the answers are spans of tokens in the articles.
We employ two different data splits by following \newcite{Zhou2017NeuralQG} \footnote{https://res.qyzhou.me/redistribute.zip} and \newcite{Du2017LearningTA} \footnote{https://github.com/xinyadu/nqg/tree/master/data/raw}. 
In \newcite{Zhou2017NeuralQG}, the original SQuAD development set is evenly divided into dev and test sets, while \newcite{Du2017LearningTA} treats SQuAD development set as its development set and splits original SQuAD training set into a training set and a test set. 
We also filter out questions which do not have any overlapped non-stop words with the corresponding sentences and perform some preprocessing steps, such as tokenization and sentence splitting. 
The data statistics are given in Table \ref{tab.dataset}.

\begin{table*}[!t]
    \centering

    \resizebox{1.0\textwidth}{!}{
    \begin{tabular}{@{}l@{~} @{~}l@{~} | @{~}c@{~} @{~}c@{~} @{~}c@{~} @{~}c@{~} @{~}c@{~} @{~}c@{~} | @{~}c@{~} @{~}c@{~} @{~}c@{~} @{~}c@{~} @{~}c@{~} @{~}c@{~}}
    \Xhline{3\arrayrulewidth}
    & & \multicolumn{6}{c}{Du Split \cite{Du2017LearningTA}} & \multicolumn{6}{c}{Zhou Split \cite{Zhou2017NeuralQG}} \\  
    \cmidrule(lr){3-8} \cmidrule(lr){9-14}
   & & B1 & B2 & B3 & B4 & MET & R-L & B1 & B2 & B3 & B4 & MET & R-L \\
    \hline
    \hline

   \multirow{2}{*}{}
  &  s2s \cite{Du2017LearningTA} & 43.09 & 25.96 & 17.50 & 12.28 & 16.62 & 39.75 & - & - & - & - & - & - \\
  & NQG++ \cite{Zhou2017NeuralQG} & - & - & - & - & - & - & - & - & - & 13.29 & - & -   \\
  & M2S+cp \cite{Song2018LeveragingCI} & - & - & - & 13.98 & 18.77 & 42.72 & - & - & - & 13.91 & - & -   \\
  & s2s+MP+GSA \cite{Zhao2018ParagraphlevelNQ} & 43.47 & 28.23 & 20.40 & 15.32 & 19.29 & 43.91 & \textbf{44.51} & 29.07 & 21.06 & 15.82 & 19.67 & 44.24 \\
  & Hybrid model   \cite{Sun2018AnswerfocusedAP} & - & - & - & - & - & - & 43.02 & 28.14 & 20.51 & 15.64 & - & - \\
  &  ASs2s \cite{Kim2019ImprovingNQ} & - & - & - & 16.20 & 19.92 & 43.96 & - & - & - & 16.17 & - & -   \\
  \hline
  & Our model & \textbf{45.66} & \textbf{30.21} & \textbf{21.82} & \textbf{16.27} & \textbf{20.36} & \textbf{44.35} & 44.40 & \textbf{29.48} & \textbf{21.54} & \textbf{16.37} & \textbf{20.68} & \textbf{44.73} \\
  
    \Xhline{3\arrayrulewidth}
    
    \end{tabular}}
    \caption{The main experimental results for our model and several baselines. `-' means no results reported in their papers. (Bn: BLEU-n, MET: METEOR, R-L: ROUGE-L) 
    }
    \label{tab:generation}
\end{table*}

We evaluate with all commonly-used metrics in question generation \cite{Du2017LearningTA}: BLEU-1 (B1), BLEU-2 (B2), BLEU-3 (B3), BLEU-4 (B4) \cite{Papineni2002BleuAM}, METEOR (MET) \cite{Denkowski2014MeteorUL} and ROUGE-L (R-L) \cite{Lin2004ROUGEAP}. 
We use the evaluation script released by \newcite{Chen2015MicrosoftCC}.

\subsection{Baseline Models} \label{sec:baseline}

We compare with the following models.
\begin{itemize}[leftmargin=*]
\setlength{\itemsep}{0pt}
\setlength{\parsep}{0pt}
\setlength{\parskip}{0pt}
    \item s2s \cite{Du2017LearningTA} proposes an attention-based sequence-to-sequence neural network for question generation.
    \item NQG++ \cite{Zhou2017NeuralQG} takes the answer position feature and linguistic features into consideration and equips the Seq2Seq model with copy mechanism.
    \item  M2S+cp \cite{Song2018LeveragingCI} conducts multi-perspective matching between the answer and the sentence to derive an answer-aware sentence representation for question generation.
    \item s2s+MP+GSA \cite{Zhao2018ParagraphlevelNQ} introduces a gated self-attention into the encoder and a maxout pointer mechanism into the decoder. We report their sentence-level results for a fair comparison.
    \item Hybrid \cite{Sun2018AnswerfocusedAP} is a hybrid model which considers the answer embedding for the question word generation and the position of context words for modeling the relative distance between the context words and the answer. 
    \item ASs2s \cite{Kim2019ImprovingNQ} replaces the answer in the sentence with a special token to avoid its appearance in the generated questions.
\end{itemize}

\subsection{Implementation Details}
We take the most frequent 20k words as our vocabulary and use the GloVe word embeddings \cite{Pennington2014GloveGV} for initialization. 
The embedding dimensions for POS, NER, answer position are set to 20.
We use two-layer LSTMs in both encoder and decoder, and the LSTMs hidden unit size is set to 600.

We use dropout \cite{Srivastava2014DropoutAS} with the probability $p=0.3$. 
All trainable parameters, except word embeddings, are randomly initialized with the Xavier uniform in $(-0.1, 0.1)$ \cite{Glorot2010UnderstandingTD}. 
For optimization in the training, we use SGD as the optimizer with a minibatch size of 64 and an initial learning rate of 1.0. We train the model for 15 epochs and start halving the learning rate after the 8th epoch. We set the gradient norm upper bound to 3 during the training.

We adopt the teacher-forcing for the training.
In the testing, we select the model with the lowest perplexity and beam search with size 3 is employed for generating questions.
All hyper-parameters and models are selected on the validation dataset.

\section{Results and Analysis}

\subsection{Main Results} \label{sec.result.main}

\begin{table*}[!h]
    \centering
    \begin{subtable}[t]{1.0\textwidth}
        \centering
        \caption{Evaluation results on all sentences.}
        \resizebox{0.75\textwidth}{!}{
        \begin{tabular}[t]{l| c c c| c c c}
            \Xhline{3\arrayrulewidth}
            \multirow{2}{*}{\textbf{}} & \multicolumn{3}{c|}{Hybrid} & \multicolumn{3}{c}{Our Model} \\ \cline{2-7}
            & BLEU & {MET} & {R-L} & {BLEU} & {MET} & {R-L} \\ 
            \hline
            0$\sim$10 ~~(72.8\% of \#) & 28.54 & 21.54 & 46.26 & 29.73 (4.17\%) & 22.03 (2.27\%) & 46.85 (1.28\%) \\
            $>$10 ~~(27.2\% of \#) & 20.67 & 16.72 & 37.63 & 22.12 (7.01\%) & 17.46 (4.43\%) &  38.47 (2.23\%) \\
            \Xhline{3\arrayrulewidth}
            \end{tabular}
            }
    \label{tab:vary_dist_a}
    \end{subtable}
    \\
    \hspace{10pt} 
    \begin{subtable}[t]{1.0\textwidth}
        \centering
        \caption{Evaluation results on sentences with more than 20 words.}
        \resizebox{0.75\textwidth}{!}{
        \begin{tabular}[t]{l| c c c| c c c}
            \Xhline{3\arrayrulewidth}
            \multirow{2}{*}{\textbf{}} & \multicolumn{3}{c|}{Hybrid} & \multicolumn{3}{c}{Our Model} \\ \cline{2-7}
            & BLEU & {MET} & {R-L} & {BLEU} & {MET} & {R-L} \\ 
            \hline
            0$\sim$10 ~~(58.3\% of \#) & 28.00 & 21.03 & 45.37 & 29.11 (3.96\%) & 21.50 (2.21\%) & 45.97 (1.31\%) \\
            $>$10 ~~(26.6\% of \#) & 20.58 & 16.66 & 37.53 & 22.04 (7.09\%) & 17.38 (4.30\%) &  38.37 (2.24\%) \\
            \Xhline{3\arrayrulewidth}
        \end{tabular}
        }
        \label{tab:vary_dist_b}
    \end{subtable}
    \caption{Performance for the average relative distance between the answer fragment and other non-stop sentence words that also appear in the ground truth question (BLEU is the average over BLEU-1 to BLEU-4). Values in parenthesis are the improvement percentage of Our Model over Hybrid. (a) is based on all sentences while (b) only considers long sentences with more than 20 words.}
    \label{tab:vary_dist}
\end{table*}

Table \ref{tab:generation} shows automatic evaluation results for our model and baselines (copied from their papers).
Our proposed model which combines structured answer-relevant relations and unstructured sentences achieves significant improvements over proximity-based answer-aware models \cite{Zhou2017NeuralQG,Sun2018AnswerfocusedAP} on both dataset splits.
Presumably, our structured answer-relevant relation is a generalization of the context explored by the proximity-based methods because they can only capture short dependencies around answer fragments while our extractions can capture both short and long dependencies given the answer fragments.
Moreover, our proposed framework is a general one to jointly leverage structured relations and unstructured sentences.
All compared baseline models which only consider unstructured sentences can be further enhanced under our framework.

Recall that existing proximity-based answer-aware models perform poorly when the distance between the answer fragment and other non-stop sentence words that also appear in the ground truth question is large (Table \ref{tab:distance}).
Here we investigate whether our proposed model using the structured answer-relevant relations can alleviate this issue or not, by conducting experiments for our model under the same setting as in Table \ref{tab:distance}.
The broken-down performances by different relative distances are shown in Table \ref{tab:vary_dist_a}.
We find that our proposed model outperforms Hybrid (our re-implemented version for this experiment) on all ranges of relative distances, which shows that the structured answer-relevant relations can capture both short and long term answer-relevant dependencies of the answer in sentences.
Furthermore, comparing the performance difference between Hybrid and our model, we find the improvements become more significant when the distance increases from ``$0\sim10$'' to ``$>10$''.
One reason is that our model can extract relations with distant dependencies to the answer, which greatly helps our model ignore the extraneous information.
Proximity-based answer-aware models may overly emphasize the neighboring words of answers and become less effective as the useful context becomes further away from the answer in the complex sentences.
In fact, the breakdown intervals in Table \ref{tab:vary_dist_a} naturally bound its sentence length, say for ``$>10$'', the sentences in this group must be longer than 10. Thus, the length variances in these two intervals could be significant. To further validate whether our model can extract long term dependency words. We rerun the analysis of Table \ref{tab:vary_dist_b} only for long sentences (length $>$ 20) of each interval. The improvement percentages over Hybrid are shown in Table \ref{tab:vary_dist_b}, which become more significant when the distance increases from ``$0\sim10$'' to ``$>10$''.

\subsection{Case Study} \label{sec.result.case}
\begin{figure}[t!]
\small
\begin{tabular}{p{0.95\columnwidth}}
\hline\hline
\textbf{Sentence}: Beyonc\'{e} received critical acclaim and commercial success, selling \hl{one million} digital copies worldwide in six days; The New York Times noted the album 's unconventional, unexpected release as significant. \\
\textbf{Reference Question}: How many digital copies of her fifth album did Beyoncé sell in six days? \\
\textbf{Baseline Prediction}: How many digital copies did the New York Times sell in six days ? \\
\textbf{Structured Answer-relevant Relation}: 
(Beyonc\'{e}; received commercial success selling; one million digital copies worldwide; in six days)\\
\textbf{Our Model Prediction}: How many digital copies did Beyonc\'{e} sell in six days ?\\
\hline
\textbf{Sentence}: The daily mean temperature in January, the area's coldest month, is 32.6 $^\circ$F (\hl{0.3 $^\circ$C}); however, temperatures usually drop to 10 $^\circ$F (-12 $^\circ$C) several times per winter and reach 50 $^\circ$F (10 $^\circ$C) several days each winter month. \\
\textbf{Reference Question}: What is New York City 's daily January mean temperature in degrees celsius ? \\
\textbf{Baseline Prediction}: What is the coldest temperature in Celsius ? \\
\textbf{Structured Answer-relevant Relation}: 
(The daily mean temperature in January; is; 32.6 $^\circ$F (0.3 $^\circ$C))\\
\textbf{Our Model Prediction}: In degrees Celsius , what is the average temperature in January ?\\
\hline\hline                
\end{tabular}
\caption{ Example questions (with answers highlighted) generated by crowd-workers (ground truth questions), the baseline model and our model.}
\label{fig.case}
\end{figure}

Figure \ref{fig.case} provides example questions generated by crowd-workers (ground truth questions), the baseline Hybrid \cite{Sun2018AnswerfocusedAP}, and our model.
In the first case, there are two subsequences in the input and the answer has no relation with the second subsequence\footnote{One might think that the two subsequences should be regarded as individual sentences, however, several off-the-shelf tools do recognize them as one sentence.}.
However, we see that the baseline model prediction copies irrelevant words ``The New York Times'' while our model can avoid using the extraneous subsequence ``The New York Times noted ...'' with the help of the structured answer-relevant relation.
Compared with the ground truth question, our model cannot capture the cross-sentence information like ``her fifth album'', where the techniques in paragraph-level QG models \cite{Zhao2018ParagraphlevelNQ} may help.
In the second case, as discussed in Section \ref{sec.intro}, this sentence contains a few facts and some facts intertwine.
We find that our model can capture distant answer-relevant dependencies such as ``mean temperature'' while the proximity-based baseline model wrongly takes neighboring words of the answer like ``coldest'' in the generated question.

\subsection{Diverse Question Generation} \label{sec.result.diverse}

Another interesting observation is that for the same answer-sentence pair, our model can generate diverse questions by taking different answer-relevant relations as input.
Such capability improves the interpretability of our model because the model is given not only what to be asked (i.e., the answer) but also the related fact (i.e., the answer-relevant relation) to be covered in the question.
In contrast, proximity-based answer-aware models can only generate one question given the sentence-answer pair regardless of how many answer-relevant relations in the sentence.
We think such capability can also validate our motivation: questions should be generated according to the answer-aware relations instead of neighboring words of answer fragments.
Figure \ref{fig.case_diverse} show two examples of diverse question generation.
In the first case, the answer fragment `Hugh L. Dryden' is the appositive to `NASA Deputy Administrator' but the subject to the following tokens `announced the Apollo program ...'.
Our framework can extract these two answer-relevant relations, and by feeding them to our model separately, we can receive two questions asking different relations with regard to the answer.

\begin{figure}[t!]
\small
\begin{tabular}{p{0.95\columnwidth}}
\hline\hline
\textbf{Sentence}: In July 1960, NASA Deputy Administrator \hl{Hugh L. Dryden} announced the Apollo program to industry representatives at a series of Space Task Group conferences. \\
\textbf{Relation 1}: (Hugh L. Dryden; [is] Deputy Administrator [of]; NASA) \\
\textbf{Question 1}: Who was the NASA Deputy Administrator in 1960 ? \\
\textbf{Relation 2}: (NASA Deputy Administrator Hugh L. Dryden; announced; the Apollo program to industry representatives at a series of Space Task Group conferences; In July 1960) \\
\textbf{Question 2}: Who announced the Apollo program to industry representatives ?\\
\hline
\textbf{Sentence}: One of the network's strike-replacement programs during that time was the game show \hl{Duel}, which premiered in December 2007. \\
\textbf{Relation 1}: (the game show Duel; premiered; in December 2007) \\
\textbf{Question 1}: What game premiered in December 2007 ? \\
\textbf{Relation 2}: (One of the network's strike-replacement programs during that time; was; the game show Duel) \\
\textbf{Question 2}: What was the name of an network 's strike - replacement programs ?\\
\hline\hline                
\end{tabular}
\caption{ Example diverse questions (with answers highlighted) generated by our model with different answer-relevant relations.}
\label{fig.case_diverse}
\end{figure}

\section{Related Work}
The topic of question generation, initially motivated for educational purposes, is tackled by designing many complex rules for specific question types \cite{Mitkov2003ComputeraidedGO,Ru2010TheFQ}. 
\newcite{Heilman2010GoodQS} improve rule-based question generation by introducing a statistical ranking model. First, they remove extraneous information in the sentence to transform it into a simpler one, which can be transformed easily into a succinct question with predefined sets of general rules. Then they adopt an overgenerate-and-rank approach to select the best candidate considering several features.

With the rise of dominant neural sequence-to-sequence learning models \cite{Sutskever2014SequenceTS}, \newcite{Du2017LearningTA} frame question generation as a sequence-to-sequence learning problem. 
Compared with rule-based approaches, neural models \cite{Yuan2017MachineCB} can generate more fluent and grammatical questions. 
However, question generation is a one-to-many sequence generation problem, i.e., several aspects can be asked given a sentence, which confuses the model during train and prevents concrete automatic evaluation.
To tackle this issue, \newcite{Zhou2017NeuralQG} propose the answer-aware question generation setting which assumes the answer is already known and acts as a contiguous span inside the input sentence.
They adopt a BIO tagging scheme to incorporate the answer position information as learned embedding features in Seq2Seq learning. 
\newcite{Song2018LeveragingCI} explicitly model the information between answer and sentence with a multi-perspective matching model.
\newcite{Kim2019ImprovingNQ} also focus on the answer information and proposed an answer-separated Seq2Seq model by masking the answer with special tokens.
All answer-aware neural models treat question generation as a one-to-one mapping problem, but existing models perform poorly for sentences with a complex structure (as shown in Table \ref{tab:distance}).

Our work is inspired by the process of extraneous information removing in \cite{Heilman2010GoodQS, cao2018faithful}. Different from \newcite{Heilman2010GoodQS} which directly use the simplified sentence for generation and \newcite{cao2018faithful} which only consider aggregate two sources of information via gated attention in summarization, we propose to combine the structured answer-relevant relation and the original sentence. 
Factoid question generation from structured text is initially investigated by \newcite{Serban2016GeneratingFQ}, but our focus here is leveraging structured inputs to help question generation over unstructured sentences. 
Our proposed model can take advantage of unstructured sentences and structured answer-relevant relations to maintain informativeness and faithfulness of generated questions.
The proposed model can also be generalized in other conditional sequence generation tasks which require multiple sources of inputs, e.g., distractor generation for multiple choice questions \cite{Gao2019GeneratingDF}. 

\section{Conclusions and Future Work}
In this paper, we propose a question generation system which combines unstructured sentences and structured answer-relevant relations for generation.
The unstructured sentences maintain the informativeness of generated questions while structured answer-relevant relations keep the faithfulness of questions.
Extensive experiments demonstrate that our proposed model achieves state-of-the-art performance across several metrics.
Furthermore, our model can generate diverse questions with different structured answer-relevant relations. For future work, there are some interesting dimensions to explore, such as difficulty levels \cite{Gao2018DifficultyCQ}, paragraph-level information \cite{Zhao2018ParagraphlevelNQ} and conversational question generation \cite{Gao2019InterconnectedQG}.

\section*{Acknowledgments}
This work is supported by the Research Grants Council of the Hong Kong Special Administrative Region, China (No. CUHK 14208815 and No. CUHK 14210717 of the General Research Fund).
We would like to thank the anonymous reviewers for their comments.
We would also like to thank Department of Computer Science and Engineering, The Chinese University of Hong Kong for the conference grant support.

\bibliography{emnlp-ijcnlp-2019} 

\begin{thebibliography}{30}
\expandafter\ifx\csname natexlab\endcsname\relax\def\natexlab#1{#1}\fi

\bibitem[{Cao et~al.(2018)Cao, Wei, Li, and Li}]{cao2018faithful}
Ziqiang Cao, Furu Wei, Wenjie Li, and Sujian Li. 2018.
\newblock Faithful to the original: Fact aware neural abstractive
  summarization.
\newblock In \emph{Thirty-Second AAAI Conference on Artificial Intelligence}.

\bibitem[{Chen et~al.(2015)Chen, Fang, Lin, Vedantam, Gupta, Doll{\'a}r, and
  Zitnick}]{Chen2015MicrosoftCC}
Xinlei Chen, Hao Fang, Tsung-Yi Lin, Ramakrishna Vedantam, Saurabh Gupta, Piotr
  Doll{\'a}r, and C.~Lawrence Zitnick. 2015.
\newblock Microsoft coco captions: Data collection and evaluation server.
\newblock \emph{CoRR}, abs/1504.00325.

\bibitem[{Denkowski and Lavie(2014)}]{Denkowski2014MeteorUL}
Michael Denkowski and Alon Lavie. 2014.
\newblock \href {https://doi.org/10.3115/v1/W14-3348} {Meteor universal:
  Language specific translation evaluation for any target language}.
\newblock In \emph{Proceedings of the Ninth Workshop on Statistical Machine
  Translation}, pages 376--380, Baltimore, Maryland, USA. Association for
  Computational Linguistics.

\bibitem[{Du et~al.(2017)Du, Shao, and Cardie}]{Du2017LearningTA}
Xinya Du, Junru Shao, and Claire Cardie. 2017.
\newblock \href {https://doi.org/10.18653/v1/P17-1123} {Learning to ask: Neural
  question generation for reading comprehension}.
\newblock In \emph{Proceedings of the 55th Annual Meeting of the Association
  for Computational Linguistics (Volume 1: Long Papers)}, pages 1342--1352,
  Vancouver, Canada. Association for Computational Linguistics.

\bibitem[{Gao et~al.(2019{\natexlab{a}})Gao, Bing, Chen, Lyu, and
  King}]{Gao2018DifficultyCQ}
Yifan Gao, Lidong Bing, Wang Chen, Michael~R. Lyu, and Irwin King.
  2019{\natexlab{a}}.
\newblock Difficulty controllable generation of reading comprehension
  questions.
\newblock In \emph{Proceedings of the Twenty-Eightth International Joint
  Conference on Artificial Intelligence, {IJCAI-19}}. International Joint
  Conferences on Artificial Intelligence Organization.

\bibitem[{Gao et~al.(2019{\natexlab{b}})Gao, Bing, Li, King, and
  Lyu}]{Gao2019GeneratingDF}
Yifan Gao, Lidong Bing, Piji Li, Irwin King, and Michael~R. Lyu.
  2019{\natexlab{b}}.
\newblock \href {https://doi.org/10.1609/aaai.v33i01.33016423} {Generating
  distractors for reading comprehension questions from real examinations}.
\newblock \emph{Proceedings of the AAAI Conference on Artificial Intelligence},
  33(01):6423--6430.

\bibitem[{Gao et~al.(2019{\natexlab{c}})Gao, Li, King, and
  Lyu}]{Gao2019InterconnectedQG}
Yifan Gao, Piji Li, Irwin King, and Michael~R. Lyu. 2019{\natexlab{c}}.
\newblock \href {https://www.aclweb.org/anthology/P19-1480} {Interconnected
  question generation with coreference alignment and conversation flow
  modeling}.
\newblock In \emph{Proceedings of the 57th Annual Meeting of the Association
  for Computational Linguistics}, pages 4853--4862, Florence, Italy.
  Association for Computational Linguistics.

\bibitem[{Glorot and Bengio(2010)}]{Glorot2010UnderstandingTD}
Xavier Glorot and Yoshua Bengio. 2010.
\newblock \href {http://proceedings.mlr.press/v9/glorot10a.html} {Understanding
  the difficulty of training deep feedforward neural networks}.
\newblock In \emph{Proceedings of the Thirteenth International Conference on
  Artificial Intelligence and Statistics}, volume~9 of \emph{Proceedings of
  Machine Learning Research}, pages 249--256, Chia Laguna Resort, Sardinia,
  Italy. PMLR.

\bibitem[{Gu et~al.(2016)Gu, Lu, Li, and Li}]{Gu2016IncorporatingCM}
Jiatao Gu, Zhengdong Lu, Hang Li, and Victor~O.K. Li. 2016.
\newblock \href {https://doi.org/10.18653/v1/P16-1154} {Incorporating copying
  mechanism in sequence-to-sequence learning}.
\newblock In \emph{Proceedings of the 54th Annual Meeting of the Association
  for Computational Linguistics (Volume 1: Long Papers)}, pages 1631--1640,
  Berlin, Germany. Association for Computational Linguistics.

\bibitem[{Gulcehre et~al.(2016)Gulcehre, Ahn, Nallapati, Zhou, and
  Bengio}]{Glehre2016PointingTU}
Caglar Gulcehre, Sungjin Ahn, Ramesh Nallapati, Bowen Zhou, and Yoshua Bengio.
  2016.
\newblock \href {https://doi.org/10.18653/v1/P16-1014} {Pointing the unknown
  words}.
\newblock In \emph{Proceedings of the 54th Annual Meeting of the Association
  for Computational Linguistics (Volume 1: Long Papers)}, pages 140--149,
  Berlin, Germany. Association for Computational Linguistics.

\bibitem[{Heilman and Smith(2010)}]{Heilman2010GoodQS}
Michael Heilman and Noah~A. Smith. 2010.
\newblock \href {https://www.aclweb.org/anthology/N10-1086} {Good question!
  statistical ranking for question generation}.
\newblock In \emph{Human Language Technologies: The 2010 Annual Conference of
  the North {A}merican Chapter of the Association for Computational
  Linguistics}, pages 609--617, Los Angeles, California. Association for
  Computational Linguistics.

\bibitem[{Kim et~al.(2019)Kim, Lee, Shin, and Jung}]{Kim2019ImprovingNQ}
Yanghoon Kim, Hwanhee Lee, Joongbo Shin, and Kyomin Jung. 2019.
\newblock Improving neural question generation using answer separation.
\newblock In \emph{AAAI Conference on Artificial Intelligence}.

\bibitem[{Lin(2004)}]{Lin2004ROUGEAP}
Chin-Yew Lin. 2004.
\newblock \href {https://www.aclweb.org/anthology/W04-1013} {{ROUGE}: A package
  for automatic evaluation of summaries}.
\newblock In \emph{Text Summarization Branches Out: Proceedings of the {ACL}-04
  Workshop}, pages 74--81, Barcelona, Spain. Association for Computational
  Linguistics.

\bibitem[{Luong et~al.(2015)Luong, Pham, and Manning}]{Luong2015EffectiveAT}
Thang Luong, Hieu Pham, and Christopher~D. Manning. 2015.
\newblock \href {https://doi.org/10.18653/v1/D15-1166} {Effective approaches to
  attention-based neural machine translation}.
\newblock In \emph{Proceedings of the 2015 Conference on Empirical Methods in
  Natural Language Processing}, pages 1412--1421, Lisbon, Portugal. Association
  for Computational Linguistics.

\bibitem[{Mitkov and Ha(2003)}]{Mitkov2003ComputeraidedGO}
Ruslan Mitkov and Le~An Ha. 2003.
\newblock \href {https://www.aclweb.org/anthology/W03-0203} {Computer-aided
  generation of multiple-choice tests}.
\newblock In \emph{Proceedings of the {HLT}-{NAACL} 03 Workshop on Building
  Educational Applications Using Natural Language Processing}, pages 17--22.

\bibitem[{Mostafazadeh et~al.(2016)Mostafazadeh, Misra, Devlin, Mitchell, He,
  and Vanderwende}]{Mostafazadeh2016GeneratingNQ}
Nasrin Mostafazadeh, Ishan Misra, Jacob Devlin, Margaret Mitchell, Xiaodong He,
  and Lucy Vanderwende. 2016.
\newblock \href {https://doi.org/10.18653/v1/P16-1170} {Generating natural
  questions about an image}.
\newblock In \emph{Proceedings of the 54th Annual Meeting of the Association
  for Computational Linguistics (Volume 1: Long Papers)}, pages 1802--1813,
  Berlin, Germany. Association for Computational Linguistics.

\bibitem[{Papineni et~al.(2002)Papineni, Roukos, Ward, and
  Zhu}]{Papineni2002BleuAM}
Kishore Papineni, Salim Roukos, Todd Ward, and Wei-Jing Zhu. 2002.
\newblock \href {https://doi.org/10.3115/1073083.1073135} {{B}leu: a method for
  automatic evaluation of machine translation}.
\newblock In \emph{Proceedings of 40th Annual Meeting of the Association for
  Computational Linguistics}, pages 311--318, Philadelphia, Pennsylvania, USA.
  Association for Computational Linguistics.

\bibitem[{Pennington et~al.(2014)Pennington, Socher, and
  Manning}]{Pennington2014GloveGV}
Jeffrey Pennington, Richard Socher, and Christopher Manning. 2014.
\newblock \href {https://doi.org/10.3115/v1/D14-1162} {{G}love: Global vectors
  for word representation}.
\newblock In \emph{Proceedings of the 2014 Conference on Empirical Methods in
  Natural Language Processing ({EMNLP})}, pages 1532--1543, Doha, Qatar.
  Association for Computational Linguistics.

\bibitem[{Rajpurkar et~al.(2016)Rajpurkar, Zhang, Lopyrev, and
  Liang}]{Rajpurkar2016SQuAD10}
Pranav Rajpurkar, Jian Zhang, Konstantin Lopyrev, and Percy Liang. 2016.
\newblock \href {https://doi.org/10.18653/v1/D16-1264} {{SQ}u{AD}: 100,000+
  questions for machine comprehension of text}.
\newblock In \emph{Proceedings of the 2016 Conference on Empirical Methods in
  Natural Language Processing}, pages 2383--2392, Austin, Texas. Association
  for Computational Linguistics.

\bibitem[{Rus et~al.(2010)Rus, Wyse, Piwek, Lintean, Stoyanchev, and
  Moldovan}]{Ru2010TheFQ}
Vasile Rus, Brendan Wyse, Paul Piwek, Mihai Lintean, Svetlana Stoyanchev, and
  Christian Moldovan. 2010.
\newblock \href {https://www.aclweb.org/anthology/W10-4234} {The first question
  generation shared task evaluation challenge}.
\newblock In \emph{Proceedings of the 6th International Natural Language
  Generation Conference}.

\bibitem[{Saha and {Mausam}(2018)}]{Saha2018OpenIE}
Swarnadeep Saha and {Mausam}. 2018.
\newblock \href {https://www.aclweb.org/anthology/C18-1194} {Open information
  extraction from conjunctive sentences}.
\newblock In \emph{Proceedings of the 27th International Conference on
  Computational Linguistics}, pages 2288--2299, Santa Fe, New Mexico, USA.
  Association for Computational Linguistics.

\bibitem[{See et~al.(2017)See, Liu, and Manning}]{See2017GetTT}
Abigail See, Peter~J. Liu, and Christopher~D. Manning. 2017.
\newblock \href {https://doi.org/10.18653/v1/P17-1099} {Get to the point:
  Summarization with pointer-generator networks}.
\newblock In \emph{Proceedings of the 55th Annual Meeting of the Association
  for Computational Linguistics (Volume 1: Long Papers)}, pages 1073--1083,
  Vancouver, Canada. Association for Computational Linguistics.

\bibitem[{Serban et~al.(2016)Serban, Garc{\'\i}a-Dur{\'a}n, Gulcehre, Ahn,
  Chandar, Courville, and Bengio}]{Serban2016GeneratingFQ}
Iulian~Vlad Serban, Alberto Garc{\'\i}a-Dur{\'a}n, Caglar Gulcehre, Sungjin
  Ahn, Sarath Chandar, Aaron Courville, and Yoshua Bengio. 2016.
\newblock \href {https://doi.org/10.18653/v1/P16-1056} {Generating factoid
  questions with recurrent neural networks: The 30{M} factoid question-answer
  corpus}.
\newblock In \emph{Proceedings of the 54th Annual Meeting of the Association
  for Computational Linguistics (Volume 1: Long Papers)}, pages 588--598,
  Berlin, Germany. Association for Computational Linguistics.

\bibitem[{Song et~al.(2018)Song, Wang, Hamza, Zhang, and
  Gildea}]{Song2018LeveragingCI}
Linfeng Song, Zhiguo Wang, Wael Hamza, Yue Zhang, and Daniel Gildea. 2018.
\newblock \href {https://doi.org/10.18653/v1/N18-2090} {Leveraging context
  information for natural question generation}.
\newblock In \emph{Proceedings of the 2018 Conference of the North {A}merican
  Chapter of the Association for Computational Linguistics: Human Language
  Technologies, Volume 2 (Short Papers)}, pages 569--574, New Orleans,
  Louisiana. Association for Computational Linguistics.

\bibitem[{Srivastava et~al.(2014)Srivastava, Hinton, Krizhevsky, Sutskever, and
  Salakhutdinov}]{Srivastava2014DropoutAS}
Nitish Srivastava, Geoffrey Hinton, Alex Krizhevsky, Ilya Sutskever, and Ruslan
  Salakhutdinov. 2014.
\newblock \href {http://jmlr.org/papers/v15/srivastava14a.html} {Dropout: A
  simple way to prevent neural networks from overfitting}.
\newblock \emph{Journal of Machine Learning Research}, 15:1929--1958.

\bibitem[{Sun et~al.(2018)Sun, Liu, Lyu, He, Ma, and
  Wang}]{Sun2018AnswerfocusedAP}
Xingwu Sun, Jing Liu, Yajuan Lyu, Wei He, Yanjun Ma, and Shi Wang. 2018.
\newblock \href {https://www.aclweb.org/anthology/D18-1427} {Answer-focused and
  position-aware neural question generation}.
\newblock In \emph{Proceedings of the 2018 Conference on Empirical Methods in
  Natural Language Processing}, pages 3930--3939, Brussels, Belgium.
  Association for Computational Linguistics.

\bibitem[{Sutskever et~al.(2014)Sutskever, Vinyals, and
  Le}]{Sutskever2014SequenceTS}
Ilya Sutskever, Oriol Vinyals, and Quoc~V Le. 2014.
\newblock \href
  {http://papers.nips.cc/paper/5346-sequence-to-sequence-learning-with-neural-networks.pdf}
  {Sequence to sequence learning with neural networks}.
\newblock In \emph{Advances in Neural Information Processing Systems 27}, pages
  3104--3112. Curran Associates, Inc.

\bibitem[{Yuan et~al.(2017)Yuan, Wang, Gulcehre, Sordoni, Bachman, Zhang,
  Subramanian, and Trischler}]{Yuan2017MachineCB}
Xingdi Yuan, Tong Wang, Caglar Gulcehre, Alessandro Sordoni, Philip Bachman,
  Saizheng Zhang, Sandeep Subramanian, and Adam Trischler. 2017.
\newblock \href {https://doi.org/10.18653/v1/W17-2603} {Machine comprehension
  by text-to-text neural question generation}.
\newblock In \emph{Proceedings of the 2nd Workshop on Representation Learning
  for {NLP}}, pages 15--25, Vancouver, Canada. Association for Computational
  Linguistics.

\bibitem[{Zhao et~al.(2018)Zhao, Ni, Ding, and Ke}]{Zhao2018ParagraphlevelNQ}
Yao Zhao, Xiaochuan Ni, Yuanyuan Ding, and Qifa Ke. 2018.
\newblock \href {https://www.aclweb.org/anthology/D18-1424} {Paragraph-level
  neural question generation with maxout pointer and gated self-attention
  networks}.
\newblock In \emph{Proceedings of the 2018 Conference on Empirical Methods in
  Natural Language Processing}, pages 3901--3910, Brussels, Belgium.
  Association for Computational Linguistics.

\bibitem[{Zhou et~al.(2017)Zhou, Yang, Wei, Tan, Bao, and
  Zhou}]{Zhou2017NeuralQG}
Qingyu Zhou, Nan Yang, Furu Wei, Chuanqi Tan, Hangbo Bao, and Ming Zhou. 2017.
\newblock Neural question generation from text: {A} preliminary study.
\newblock In \emph{Proceedings of the 6th {CCF} International Conference on
  Natural Language Processing and Chinese Computing ({NLPCC})}, pages 662--671,
  Dalian, China.

\end{thebibliography}
\bibliographystyle{acl_natbib}

\end{document}